\documentclass{article}

\usepackage[preprint]{neurips_2021_imagenet}

\usepackage[utf8]{inputenc} 
\usepackage[T1]{fontenc}    
\usepackage{hyperref}       
\usepackage{url}            
\usepackage{booktabs}       
\usepackage{amsfonts}       
\usepackage{nicefrac}       
\usepackage{microtype}      
\usepackage{xcolor}         
\usepackage{graphicx}
\usepackage{floatrow}
\usepackage{gensymb}

\title{ShapeY: Measuring Shape Recognition Capacity Using Nearest Neighbor Matching}

\author{
  Jong Woo Nam\\
  Neuroscience Graduate Program\\
  University of Southern California\\
  Los Angeles, CA, 90007\\
  \texttt{namj@usc.edu} \\
   \And
   Amanda S. Rios \\
   Neuroscience Graduate Program \\
   University of Southern California\\
   Los Angeles, CA, 90007\\
   \texttt{amandari@usc.edu} \\
   \And
   Bartlett W. Mel \\
   Biomedical Engineering Department\\
   University of Southern California\\
   Los Angeles, CA, 90007\\
   \texttt{mel@usc.edu} \\
}

\begin{document}

\maketitle

\begin{abstract}
Object recognition in humans depends primarily on shape cues. We have developed a new approach to measuring the shape recognition performance of a vision system based on nearest neighbor view matching within the system's embedding space. Our performance benchmark, ShapeY, allows for precise control of task difficulty, by enforcing that view matching span a specified degree of 3D viewpoint change and/or appearance change. As a first test case we measured the performance of ResNet50 pre-trained on ImageNet. Matching error rates were high. For example, a 27\degree change in object pitch led ResNet50 to match the incorrect object 45\% of the time. Appearance changes were also highly disruptive.  Examination of false matches indicates that ResNet50's embedding space is severely "tangled". These findings suggest ShapeY can be a useful tool for charting the progress of artificial vision systems towards human-level shape recognition capabilities.

\end{abstract}

\section{Introduction}
Object recognition in humans is based primarily on shape, \citep{grill-spector_lateral_2001,biederman_recognition-by-components_1987,biederman_surface_1988, hoffman_visual_1998, kourtzi_representation_2001}.  In contrast, deep networks (DNs) trained on conventional object and scene datasets such as ImageNet have only a weak grasp of true shape, instead classifying images mainly based on color, texture, local shape, and context cues \citep{baker_deep_2018, brendel_approximating_2019, geirhos_imagenet-trained_2019}.  The lack of a "shape bias" in conventional DNs is a likely contributor to the various un-biological performance characteristics of DNs, including their susceptibility to adversarial inputs \citep{goodfellow_explaining_2015}; their propensity to confidently classify random noise patterns as specific objects; their poor generalization behavior; and their inability to recognize objects based on line drawings, though line drawings explicitly represent the key information needed for recognition in humans \citep{brendel_approximating_2019, geirhos_imagenet-trained_2019, russakovsky_imagenet_2015}.

A poor grasp of global object shape does not prevent conventional DNs from performing well on benchmark tasks, however: state-of-the-art top-5 performance on ImageNet is approaching 99\% \citep{pham_meta_2021}.  It therefore seems that ImageNet does not test, and is evidently not well suited for training, the basic representational capability that underlies human object and scene vision.

We have developed a new benchmarking approach based on the idea – similar to that motivating some contrastive learning approaches  \citep{chen_simple_2020} – that the core competence of a 3D recognition system is to produce similar internal visual codes when familiar objects or scenes are viewed from different perspectives, under different lighting conditions, and/or with different backgrounds.  On the other hand, to state the obvious, the recognizing system should produce different codes for different objects, regardless of viewpoint, lighting conditions, etc.  
The ability to perform well at this basic matching task is, in our view, a pre-requisite for performing well and generalizing well in real-world object recognition tasks.  Conversely, a recognition system that performs poorly at this task cannot be said to understand shape.
In the following, we describe our image set, our measures of task performance, and the way we control task difficulty.

\begin{figure}
    \centering
    \includegraphics[width=12cm]{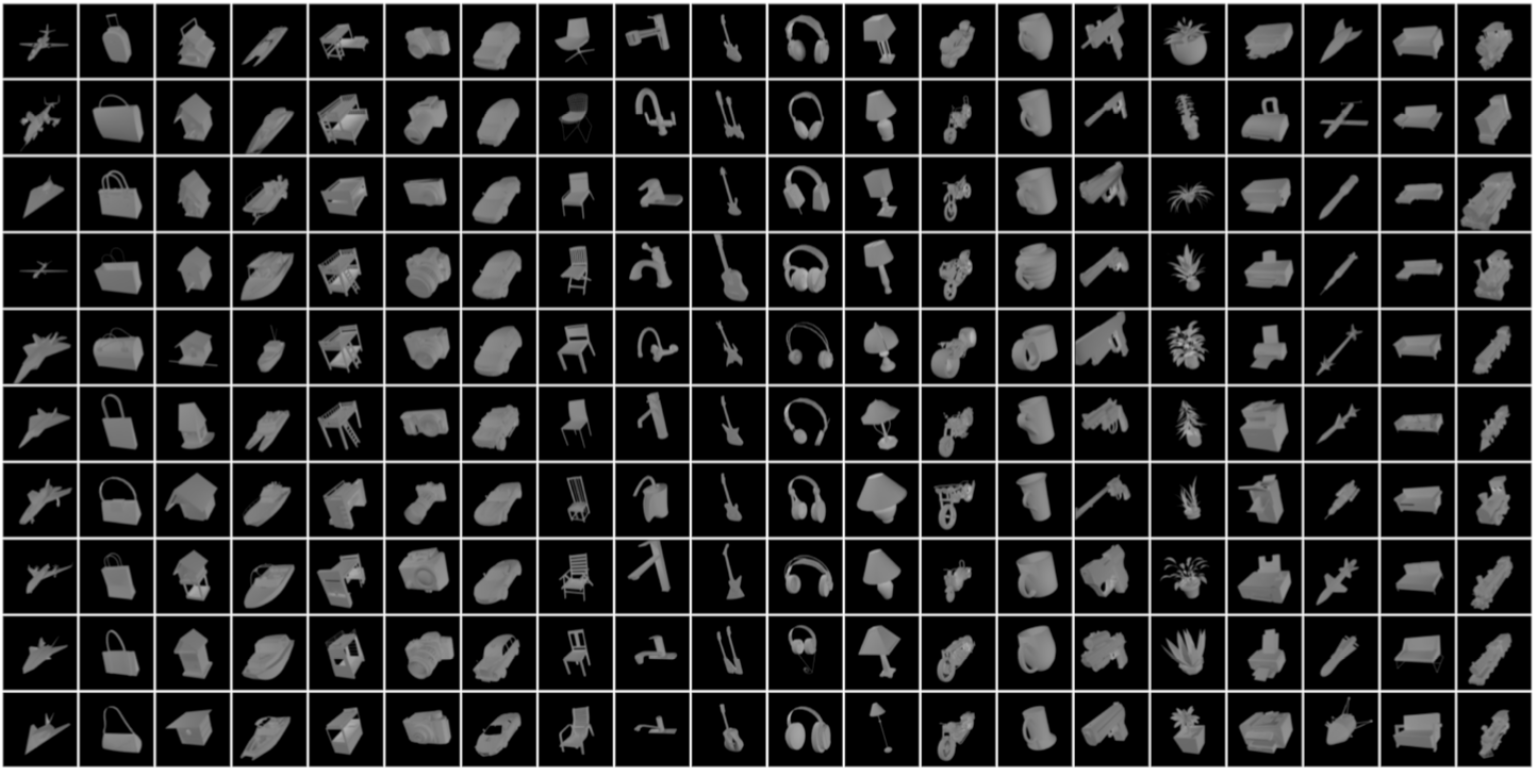}
    \caption{Complete set of 3D object models (from shapenet.org) rendered using Blender at their "origin viewpoints". Objects were grouped into 20 categories, with 10 instances of each.}
    \label{fig:all_objects}
\end{figure}

\section{The ShapeY Image Set}
Constructing a challenging 3D shape recognition benchmark based on nearest-neighbor view matching requires (1) the ability to densely sample 3D views for each represented object; (2) the ability to manipulate (or eliminate) all non-shape-related cues; and (3) the availability of a large and diverse set of 3D object models. \citet{borji_ilab-20m_2016} provide a comprehensive review of 3D view-based image databases; some meet one or two of these requirements, but none – or any other database that we are aware of – satisfies all three.  We therefore set out to produce a new object view database using Blender and publicly available 3D models from Shapenet.org \citep{chang_shapenet_2015}.

Our image database currently contains $\sim$ 63,000 rendered images of 3D objects. Each 256x256 image depicts a single object, grey in color, against a black background.  The database contains 20 basic level object categories (chair, airplane, plant, etc.), 10 instances of each category (Figure \ref{fig:all_objects}), and 321 3D views of each instance. Object views are grouped into "series" representing different combinations of viewpoint transformations (CVTs). Each series is centered on a common "origin view" of the object, with 5 viewpoint steps moving away from the origin in both directions for a total of 11 views per series.  Five types of rigid transformation were used (x, y, pitch, roll and yaw; scale changes were excluded so as to preserve object detail that would be lost at smaller scales), leading to 31 possible CVTs (31 = 5 transformations chosen 1, 2, 3, 4, or 5 at a time). In each viewpoint step the object was transformed simultaneously along all dimensions in the CVT. For example, in a series combining "x" and "roll", each step in the series came with a horizontal shift of $\sim$3.3\% of the image width, combined with 9\degree of image plane rotation.  In series’ containing "pitch" or "yaw", the object was rotated in depth around the horizontal or the vertical axis of the object, respectively.  The step sizes meant that over the 10 steps from one end of a series to the other, the object could shift by $\sim$33\% of the image width and/or rotate by 90\degree, or both.  Examples of all series that contained  transformations in both pitch and yaw ('pw') are shown in Figure 2.

\begin{figure}
    \centering
    \floatbox[{\capbeside\thisfloatsetup{capbesideposition={right,top},capbesidewidth=4.5cm}}]{figure}[\FBwidth]
    {\caption{Positive match candidates (PMCs) for view \#8 (blue box) out of 11 in the series with CVT = 'pw'. Rows show all 8 series containing 'pw'. The difficulty of the matching task is controlled by excluding positive match candidates in the "vicinity" of the reference view in viewpoint space. The "exclusion zone" shown (red shading) is for an exclusion radius $r_e=2$.} \label{fig:transforms_and_exclusions}}
    {\includegraphics[height=2.5in]{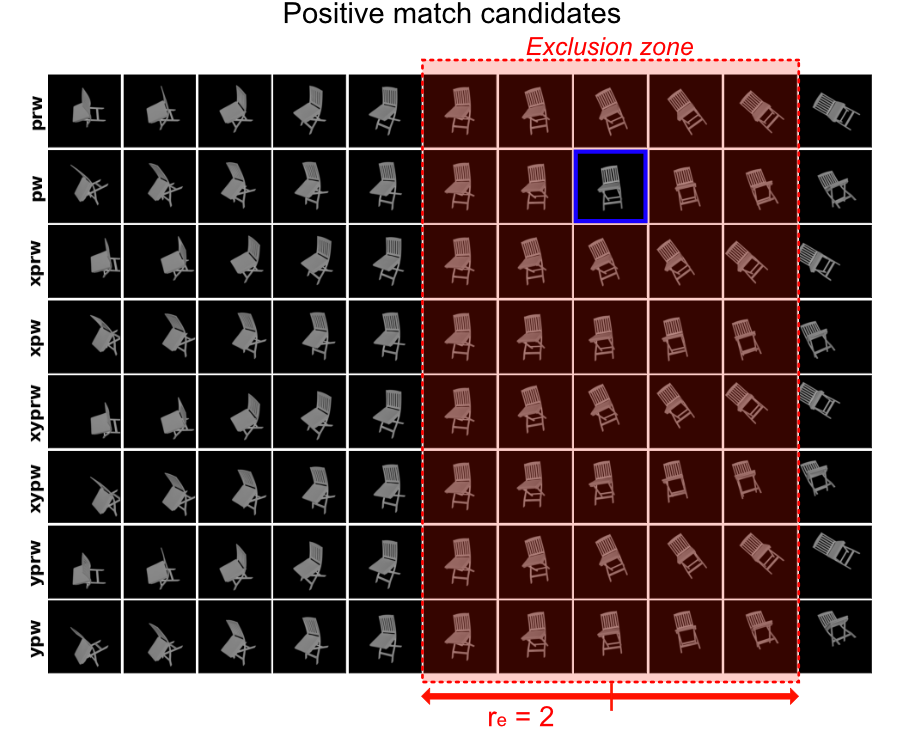}}
\end{figure}

\section{Novel features of the ShapeY performance benchmark}

ShapeY has three notable features. First, it is designed to probe the micro-structure of the embedding space of a shape-representing network by directly asking "what looks most similar to what" in that space. Given an input image, a response is scored as "correct" if the closest match is to another view of the same object; "categorically correct" if the closest match is to a view of a different object within the same category; and "incorrect" if the closest match is to a "distractor" from a different object category.  Thus, unlike most OR benchmarks in wide use, in which a response is considered correct if an object view is rated as \emph{generally} better matched to the correct class compared to other classes (e.g. by computing cosine distance to a class prototype in the final layer of a DN), our benchmark enforces the stronger condition that there should be \emph{no single view of any other object} that matches an input better than the best-matching same-object view. Thus, given a particular view of a lamp, even if 99 out of 100 of the closest views in the database are images of the same lamp, if the single closest match is a view of a boat, the trial is scored as an error.  Measuring performance in this "worst case" manner allows us to more sensitively detect "tangles" in the fine structure of the shape-space embedding \citep{dicarlo_untangling_2007}.

Second, our benchmark allows task difficulty to be finely controlled through the use of "exclusions". In the case of a viewpoint exclusion, we choose an exclusion radius $r_e$, and then eliminate as positive match candidates (PMCs) all same-object views surrounding, and therefore most similar to, the input view, up to $r_e$ steps along a designated set of transformation dimensions.  For example, if $r_e=2$, and 'pw' is the designated set of exclusion transformations, then PMCs must be at least 3 viewpoint steps away from the reference view in both pitch and yaw, and can therefore be drawn only from the 8 series whose CVTs contain both pitch and yaw ('pw', 'xpw', 'xypw', 'xprw', 'xyprw', 'ypw', 'prw', 'yprw') (Figure \ref{fig:transforms_and_exclusions}). This particular set of exclusion parameters guarantees that any successful match to a reference view must have bridged at least a 27\degree change in both pitch and yaw, and could also differ from the reference view by 3 viewpoint steps along one or more other viewpoint dimensions. Measuring the decline in matching performance as $r_e$ increases allows us to quantify the degree of 3D viewpoint variation that the shape-representing system can tolerate before false matches to similar-looking distractor objects begin to increase in frequency. Similarly, by varying the composition of the exclusion transformation set, we can test which dimensions of viewpoint variation, singly or in combination, are most disruptive to shape-matching performance.

In addition to ignoring modest changes in viewpoint, a shape-based recognition system must be capable of ignoring changes in non-shape cues, including the colors and textures of objects and backgrounds, changes in lighting conditions, etc. (this is the main idea underlying contrastive learning). We quantified the ability to cope with these types of changes through the use of "appearance exclusions".  An example of an appearance exclusion would be a "contrast exclusion", in which object views rendered in the original format with black backgrounds (Figure \ref{fig:all_objects}) could only be matched to views of themselves with light backgrounds. That is, views containing the original black backgrounds were excluded from the set of PMCs. To put this into practice, we doubled the database to include a second, light background version of every object view. In the "hard" version of the contrast exclusion task, given a "reference" view, all same-object views with dark backgrounds were excluded as match candidates, forcing the system to recognize the same shape despite the change in background. All views of \emph{other} objects were not subject to the exclusion, however, and were available to match in the original black background only. In the "soft" version of the task, \emph{all} match candidates were subject to the exclusion, including all views of different objects and of the same object.  

\begin{figure}
    \centering
    \floatbox[{\capbeside\thisfloatsetup{capbesideposition={right,top},capbesidewidth=3.3cm}}]{figure}[\FBwidth]
    {\caption{Nearest neighbor matching error, plotted against the exclusion radius $r_e$. Top row shows errors for object matching; bottom row shows errors for category matching. Three columns show results for exclusion transformation sets with 1, 2, and 3 viewpoint transformation dimensions, respectively.} \label{fig:nnerror_graphs}}
    {\includegraphics[height=8.3cm]{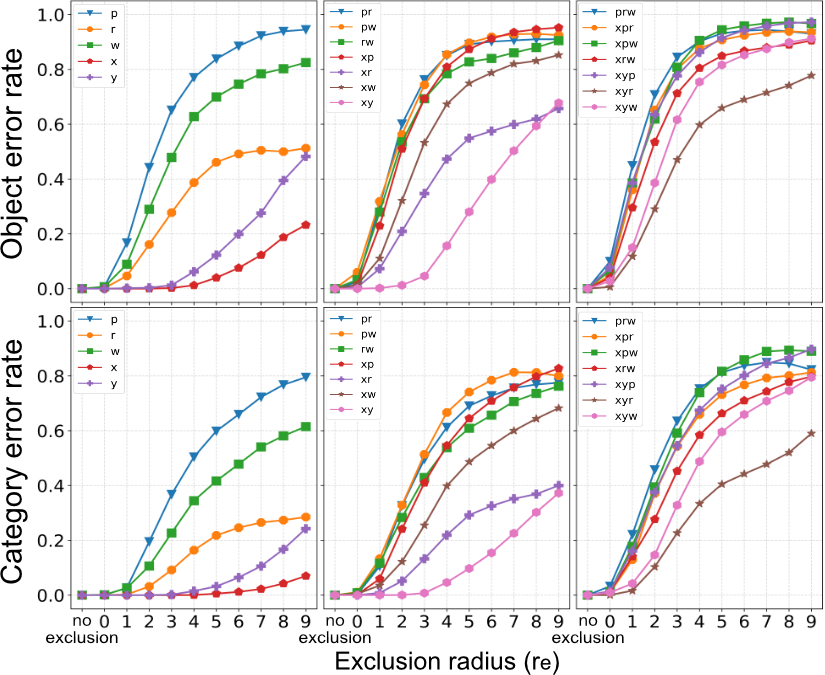}}
\end{figure}

\section{Results}

We tested the performance of a ResNet50 \citep{he_deep_2015}, pre-trained on ImageNet \citep{paszke_pytorch_2019}. Results of a basic matching test are shown in Figure \ref{fig:nnerror_graphs}. Error rates averaged over all 63,000 views are shown in Figure \ref{fig:nnerror_graphs} for all exclusion transformations involving either 1, 2, or 3 transformation dimensions (columns 1-3, respectively). 
Error rates were surprisingly high.
For the single transformation dimension 'p', error rate was already 45\% for $r_e$= 2, corresponding to an enforced 27\degree change in object pitch.  When pitch and yaw were combined ('pw'), the error rate climbed to nearly 60\% at $r_e$= 2. Error rates were generally worse when more transformation dimensions were combined; worse for depth rotations than image plane rotations; and much worse for rotations than shifts. (The near perfect shift invariance for $r_e=2$ was expected given the ResNet50 embedding was taken from the global average pooling layer, which explicitly pools across image shifts).  Category error rates were lower, but remained substantial. For example, a 27\degree change in object pitch and roll led to a 33\% error rate, meaning that 1 in every 3 views in the database was judged to be most similar to view from an entirely different object category.

\begin{figure}
    \centering
    \floatbox[{\capbeside\thisfloatsetup{capbesideposition={left,bottom},capbesidewidth=5cm}}]{figure}[\FBwidth]
    {\caption{Eight examples of nearest-neighbor matching errors by a pre-trained ResNet50. Within each group of 3 images, the best match to the reference view (according to ResNet50) is shown at left, and a more similar appearing correct match rejected by the DN is shown at right.  Values in orange show the correlation between that view's embedding vector and that of the reference view.\newline} \label{fig:error_examples}}
    {\includegraphics[height=5cm]{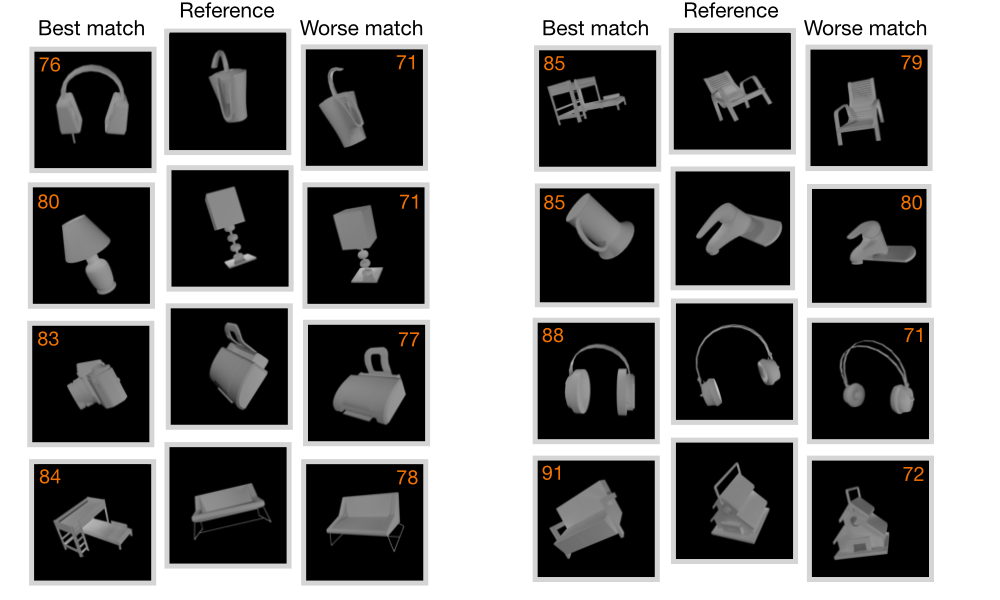}}
\end{figure}

The numerical results shown in Figure \ref{fig:nnerror_graphs} provide a quantitative summary of the shape representation capabilities of a vision system, and especially the ability to tolerate 3D viewpoint variation.  Our approach to nearest-neighbor matching with exclusions can also provide a qualitative measure of the "tangledness" of the shape embedding by analyzing shape match failures.
Match failures are particularly informative regarding the quality of the shape embedding: when a view of a reference object is found to be very close to a view of even one other object of very different shape, it is likely that the reference view is close to a large number of other very different shapes as well, whose discovery depends mainly on having a sufficient number of distractors in the view database.  Several examples of match failures are shown in Figure \ref{fig:error_examples}, along with same-object match candidates that were rejected by the DN.

\begin{figure}[h]
    \centering
    \floatbox[{\capbeside\thisfloatsetup{capbesideposition={right,top},capbesidewidth=6cm}}]{figure}[\FBwidth]
    {\caption{Nearest-neighbor matching errors when a "contrast exclusion" was compounded with viewpoint exclusions. Results shown is for the exclusion transformation set 'pr'. The first point on the x-axis means that \emph{all} contrast-reversed views, including the reference view itself, were available as positive match candidates.} \label{fig:contrast_reversal}}
    {\includegraphics[height=5cm]{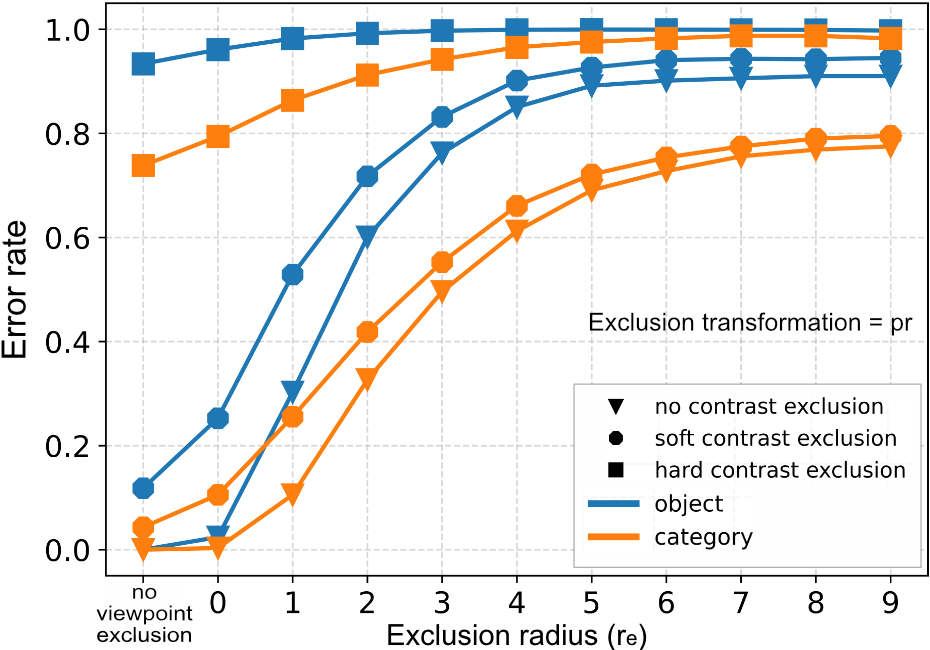}}
\end{figure}

We next tested matching performance of ResNet50 with the added challenge of a "contrast exclusion".  Figure \ref{fig:contrast_reversal} shows results for the exclusion transformation 'pr' using the both object and category matching criteria. When a reference view could only be matched to contrast reversed views (soft version), category error rates climbed from 33\% to 42\% at $r_e$= 2, and object matching errors climbed above 70\%. The difficulty encountered by ResNet50 in matching contrast reversed views is striking: even when the PMCs for a reference view included \emph{the exact same object view} but for the change in background, the reference view was falsely matched more than 70\% of the time to an object from an entirely different category.  

\section{Discussion}


As an approach to OR performance testing, pairwise view matching has three particular merits. First, the ability to recognize familiar objects and scenes despite modest changes in viewing distance, angle, lighting, and background is an innate capability of biological vision systems, and is arguably a more fundamental visual capability than image classification. Thus, we can imagine a vision system that can reliably rate the similarity of two views of an object without knowing the object class, but it would seem paradoxical if a system could correctly classify objects, while being unable to rate the similarity of two object views. In short,  pairwise view matching  is a good starting point for evaluating recognition competence. Second, our approach allows task difficulty to be controlled by parametrically varying the set of views that are qualified to be positive match candidates for any given reference view. Third, collecting and analyzing matching errors allows us to draw a rough equivalence between an amount of viewpoint change, which should minimally alter an object's shape code, and an amount of actual shape change. If we discover that a modest depth rotation of an object regularly alters the embedding vector as much as switching to a different object category, then the embedding is performing poorly. This applies to ResNet50 when pre-trained on ImageNet: a viewpoint change of 27\degree in pitch and yaw alters the embedding vector as much as switching from a birdhouse to a couch, or from a faucet to a beer mug (Figure \ref{fig:contrast_reversal}).  Likewise, if switching the background of an object from dark to light changes the embedding vector as much as a change of object category, we may conclude the embedding space is badly entangled. This can serve as a cautionary note when utilizing embeddings on downstream transfer learning tasks.

It is worth noting that, when views of objects of very different shape are found to lie near to each other in the embedding space, then it is likely that every object view is near to a panoply of different shapes. Therefore as the number of different objects in the database increases, and the embedding space becomes more densely populated with views, the rate of false matches is likely to approach 100\%.  As a reference point, our database currently contains 20 basic level object categories; by comparison, the number of basic level categories that a human subject effortlessly commands is $\sim$100-fold larger (in the range of 1-$\sim$3,000 \citep{biederman_recognition-by-components_1987} see Footnote 10).


\bibliography{references}




\end{document}